\documentclass[journal]{IEEEtran}
\usepackage{graphicx}
\usepackage{color,soul}
\usepackage{amsmath, amssymb}
\usepackage{algorithm}
\usepackage{booktabs} % For better table formatting
\usepackage{multirow} 
\usepackage{orcidlink}
\usepackage{hyperref}
\ifCLASSINFOpdf
\else
\fi
\usepackage{algorithmic}
\begin{document}
\title{Data-Driven Contact-Aware Control Method for Real-Time Deformable Tool Manipulation: A Case Study in the  Environmental Swabbing}

\author{Siavash Mahmoudi, Amirreza Davar, Dongyi Wang
%\thanks{}% <-this % stops a space
\thanks{S. Mahmoudi, A. Davar and D. Wang are with the University of Arkansas, Fayetteville, AR, USA, 72701}}% <-this % stops a space


\markboth{Journal of \LaTeX\ Class Files,~Vol.~14, No.~8, August~2021}%
{Shell \MakeLowercase{\textit{et al.}}: A Sample Article Using IEEEtran.cls for IEEE Journals}



\maketitle
\begin{abstract}
Deformable Object Manipulation (DOM) remains a critical challenge in robotics due to the complexities of developing suitable model-based control strategies. Deformable Tool Manipulation (DTM) further complicates this task by introducing additional uncertainties between the robot and its environment. While humans effortlessly manipulate deformable tools using touch and experience, robotic systems struggle to maintain stability and precision. To address these challenges, we present a novel State-Adaptive Koopman LQR (SA-KLQR) control framework for real-time deformable tool manipulation, demonstrated through a case study in environmental swab sampling for food safety. This method leverages Koopman operator-based control to linearize nonlinear dynamics while adapting to state-dependent variations in tool deformation and contact forces. A tactile-based feedback system dynamically estimates and regulates the swab tool’s angle, contact pressure, and surface coverage, ensuring compliance with food safety standards. Additionally, a sensor-embedded contact pad monitors force distribution to mitigate tool pivoting and deformation, improving stability during dynamic interactions. Experimental results validate the SA-KLQR approach, demonstrating accurate contact angle estimation, robust trajectory tracking, and reliable force regulation. The proposed framework enhances precision, adaptability, and real-time control in deformable tool manipulation, bridging the gap between data-driven learning and optimal control in robotic interaction tasks.
\end{abstract}
\bigskip % Adds vertical space between the abstract and the Note to Practitioners

\noindent\textbf{\textit{Note to Practitioners}— This study introduces a novel robotic solution to automate environmental swab sampling in food processing environments, currently a task subject to human uncertainties. By leveraging State-Adaptive Koopman Linear Quadratic Regulator (SA-KLQR), we address the challenge of manipulating deformable tools like swab sticks, which are essential for ensuring food safety. Our tactile-based feedback system embedded within the swab tool dynamically adjusts for optimal contact pressure and angle, significantly enhancing the precision and consistency of sample collection.
The embedded sensors monitor force distribution in real time, allowing the robot to adapt to changes in tool stiffness and wetness as the swab absorbs liquids. This adaptation is critical as it ensures comprehensive surface coverage and maintains strict adherence to food safety standards, mitigating risks associated with manual sampling errors.
Our approach not only reduces the reliance on human operators for repetitive and delicate tasks but also improves the reliability of the sampling process. While promising, the current system requires further refinement in sensor technology and control algorithms to enhance its effectiveness. Future developments will focus on integrating more advanced machine learning models to extend the system's adaptability and response to environmental variations.
This implementation could serve as a scalable solution that can be integrated into existing robotic systems within the industry, potentially transforming food safety protocols by automating a critical aspect of food processing and quality control. }.
\bigskip


% \fi
\IEEEpeerreviewmaketitle

\section{INTRODUCTION}
\IEEEPARstart{A}{s} automation advances, robots are increasingly utilized for complex tasks, reducing manual labor in hazardous environments while improving efficiency, precision, and cost-effectiveness \cite{9721534}. However, real-world robotic applications require seamless interaction with deformable objects, which presents significant challenges due to material flexibility and unpredictable shape changes \cite{yin2021modeling}. Unlike rigid object manipulation, deformable object manipulation (DOM) requires real-time adaptive control to compensate for continuous state variations and external forces.

Traditional physics-based control models, such as mass-spring systems and finite element methods \cite{10850721, kaufmann2009flexible, cretu2008neural}, attempt to model deformable object behavior but often fall short in real-world applications due to the sensitvity of control parameters and the difficulty of modeling complex contact dynamics. To address these limitations, recent research has shifted toward machine learning and data-driven approaches, where robots learn from sensor feedback or demonstrations rather than relying on hard-coded models \cite{yu2022shape}. Predictive learning models \cite{zhaole2024dexdlo, 9813374, 9810841} have proven effective for latent space learning and object behavior forecasting, improving adaptability across applications such as fabric repositioning \cite{wang2024efficient}, crop harvesting \cite{visentin2023soft, li2019factors}, medical robotics \cite{liu2021real}, and deformable linear object manipulation \cite{10093017, laezza2021reform}.

\begin{figure}
    \centering
    \includegraphics[width=1\linewidth]{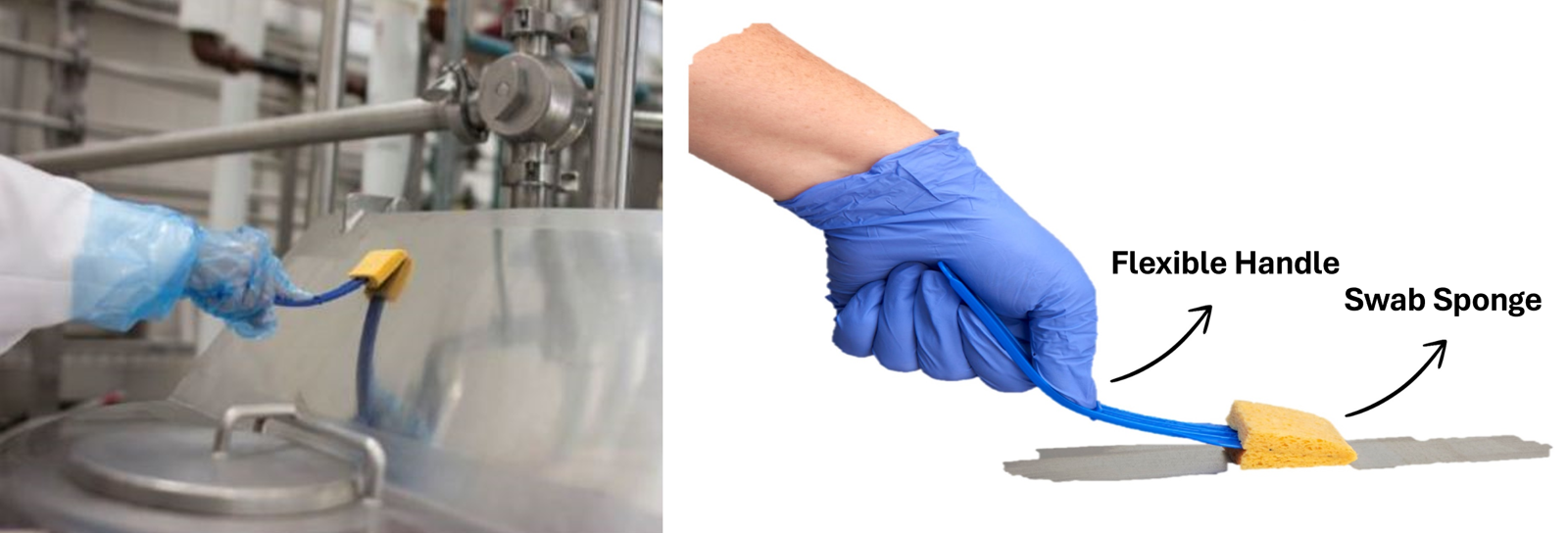}
    \caption{Swab sampling collection model with deformable sponge stick}
    \label{fig:1}
\end{figure}

While significant progress has been made in DOM, little research has focused on deformable tool manipulation (DTM), which introduces additional complexities such as bending dynamics, force regulation, and stability issues. Unlike standard soft objects, deformable tools interact with both the robot and the environment, making control significantly more challenging. This problem is particularly relevant in environmental swab sampling for food safety, where a deformable tool with a flexible handle and soft sponge need to be manipulated to collect environmental swabbing samples under strict safety protocols to protect food safety in food processing plant and the entire supply chain. This is a time consuming process for trained worker to collect the sample in large scale plants, and there is no research reported to use the robot to automate the process. A similar study reported a robotic nasopharyngeal swabbing system for COVID-19 testing \cite{shen2020robots}, while environmental swabbing in food processing plants introduces additional technical challenges. Unlike the small, predictable, and uniform anatomical cavity in nasopharyngeal applications, industrial surfaces exhibit significant variations in shape, texture, and contamination levels, necessitating continuous force adaptation and precise surface coverage. Furthermore, during the swabbing process, the swab stick itself undergoes dynamic physical changes, especially its wetness, which alters its stiffness, weight, and frictional properties in real time. 

Recent advancements in reinforcement learning (RL) and imitation learning (IL) have enhanced robotic manipulation, enabling learning-based controllers that can generalize across tasks without explicitly modeling system dynamics \cite{mahmoudi2024leveraging}. Techniques such as Behavior Cloning (BC) and adversarial IL \cite{han2023survey} allow robots to learn directly from human demonstrations. However, accurate IL performance heavily depends on high-quality demonstrations, and in cases of DTM, like swab sampling, where contact rich force control is important to the success of the task. While in practice even human experts struggle to achieve satisfactory performance such as
maintaining consistent force and swabbing coverage during the operation, and learning from such human demonstrations may introduce systematic errors \cite{9927439}. This limitation underscores the need for a model-driven ground truth to provide accurate force and trajectory references in DTM, which can also potentially benefit IL training.

This research builds on techniques from ear surgical devices \cite{8322280} and nasal swab testing \cite{10070885}, where precise force control is crucial. However, unlike these applications—where the environment or object is deformable—this study focuses on a deformable tool, introducing distinct challenges. As shown in Fig. \ref{fig:1}, the flexible handle and wet soft sponge complicate force regulation, making comprehensive sample collection more difficult.
This paper presents a novel tactile-based robotic swab sampling platform, integrating sensor feedback and Koopman operator-based optimal control to regulate force and trajectory in real-time deformable tool manipulation. The proposed approach consists of two key components: (i) an embedded waterproof contact sensor inserted inside the sponge provides real-time force feedback, allowing the robot to adjust its handle bending for optimal contact force, and (ii) once the force is stabilized, the robot follows a zigzag trajectory to ensure complete surface coverage following the food industrial standard.

To address the nonlinear behavior of the force sensor and deformable tool handle and sponge, we propose State-Adaptive Koopman Linear Quadratic Regulator (SA-KLQR), a novel Koopman-based control model that dynamically adapts force regulation based on state variations. Unlike traditional approaches, SA-KLQR switches between multiple Koopman operators based on real-time system conditions, allowing for robust adaptation to varying contact forces and handling deformations. Additionally, a sensor-embedded contact pad monitors force distribution across the contact area, enabling the system to dynamically correct tool pivoting and deformation, ensuring stability throughout the swabbing process.

To the best of our knowledge, this is one of the first tactile-based approaches for deformable tool manipulation, providing a structured framework that bridges data-driven learning with optimal control. Our method not only improves real-time robotic force regulation but also lays the groundwork for future integration with imitation learning models, enabling precise and adaptive manipulation without reliance on external force sensors. The remainder of this paper details the system design, control framework, and experimental validation, demonstrating the feasibility and effectiveness of the proposed approach. The video and code are available on the \href{https://siamo-arch.github.io/SA-KLQR/#-s}{project website}.

\section{RELATED WORK}

Force control in robotic manipulation has been extensively studied, with approaches such as hybrid force-motion control \cite{ding2024task}, impedance and admittance control \cite{bai2021spherical}, and model predictive control (MPC) \cite{gold2022model}. These methods have demonstrated effectiveness in structured environments but often struggle in dynamic and deformable contact interactions, particularly when dealing with unmodeled nonlinearities. Recent advancements in data-driven force estimation and Koopman operator-based system identification have enabled these controllers to adaptively model nonlinear dynamics, improving force regulation and stability in complex contact tasks \cite{gu2023survey, zhao2024kalman}.

Various studies have attempted to address force control challenges in deformable and soft object manipulation by developing model-based approaches that approximate interaction dynamics for control design. In \cite{wen2020force}, an admittance-based force regulation controller utilizes sEMG-based force prediction for precise gripping force control of fragile and deformable objects. \cite{kumar2021sliding} introduces a sliding-mode control algorithm for soft robotic fingers, ensuring stable in-hand manipulation by regulating internal forces without requiring object shape information. Similarly, \cite{xia2024viscoelastic} proposes a viscoelastic model-based force control framework for orthopedic surgical robots, integrating tissue model identification, preoperative force optimization, and a secure pre-touch strategy to enhance precision and safety in soft tissue interactions. While these methods manage interaction nonlinearity through model linearization techniques such as feedback linearization and dynamic inversion, or nonlinear control strategies like sliding mode control (SMC) and adaptive impedance control, their nonlinear and non-invertible structure limits compatibility with conventional model-based control techniques, restricting their direct integration into standard optimal control frameworks..

To address these limitations, Koopman operator theory has emerged as a promising solution, offering a linear representation of nonlinear dynamics that enhances compatibility with model-based control techniques \cite{brunton2021modern}. By lifting system states into a higher-dimensional space, Koopman-based methods enable optimal control strategies, allowing the application of linear control tools to inherently solve nonlinear systems dynamics. This approach has demonstrated superior performance in contact-rich tasks, where conventional controllers struggle with system uncertainties and dynamic interactions \cite{williams2016extending, proctor2018generalizing, brunton2016koopman}. In particular, recent studies have successfully applied Koopman operators to force regulation, improving tracking accuracy, stability, and robustness in robotic interaction control from manipulation to locomotion \cite{shi2024koopman}. The approach in \cite{bruder2020data} demonstrated that a Koopman-based MPC controller significantly improved trajectory tracking in a pneumatic soft robot arm, achieving more than three times higher precision compared to traditional methods. Expanding on this, , demonstrated on a quadcopter and real robotic systems, \cite{abraham2019active} presents an active learning strategy for Koopman-based control, where robotic systems autonomously refine their Koopman dynamics using information-theoretic methods, enabling faster model learning, improved LQ control synthesis, and real-time stabilization. Further improving Koopman-based control, \cite{mamakoukas2021derivative} introduces a derivative-based Koopman framework for real-time nonlinear system identification and control, using Taylor series error bounds for model accuracy. Validated on an inverted pendulum and a tail-actuated robotic fish, it outperforms Sparse Identification of Nonlinear Dynamical Systems (SINDy) and Nonlinear Autoregressive Network with Exogenous Inputs (NARX) in adaptive control under disturbances.

While Koopman-based modeling offers linear control synthesis for nonlinear dynamics, its reliance on finite-dimensional approximations introduces model bias and offline errors, limiting adaptability to real-time force variations. The observable selection remains non-trivial, often requiring trial and error, and Koopman models struggle with discontinuous or hybrid dynamics, such as deformable contact interactions in robotic swabbing \cite{ren2022koopman}. Additionally, Koopman models are typically computed offline, making them less responsive to real-time force disturbances, which is crucial in DTM force control, specially for maintaining consistent force distribution of swabbing sponge across a surface in this research application.

Despite the aforementioned limitations, Koopman operators provide a structured, data-driven alternative to black-box models, making them suitable for force-controlled swabbing by enabling efficient force regulation while maintaining computational feasibility.  In this research, by adding an external contact sensor and integrating a newly proposed centroid-based force algorithm, the SA-KLQR model can mitigate offline modeling bias, improving real-time adaptability and stability. This hybrid approach enhances force control precision, reduces deformation-induced errors, and ensures stable force control, making Koopman-based modeling a viable solution for DTM.
%\IEEEpubidadjcol

\section{SYSTEM AND FRAMEWORK IDENTIFICATION}
\begin{figure} [h]
\centering \includegraphics[width=1\linewidth]{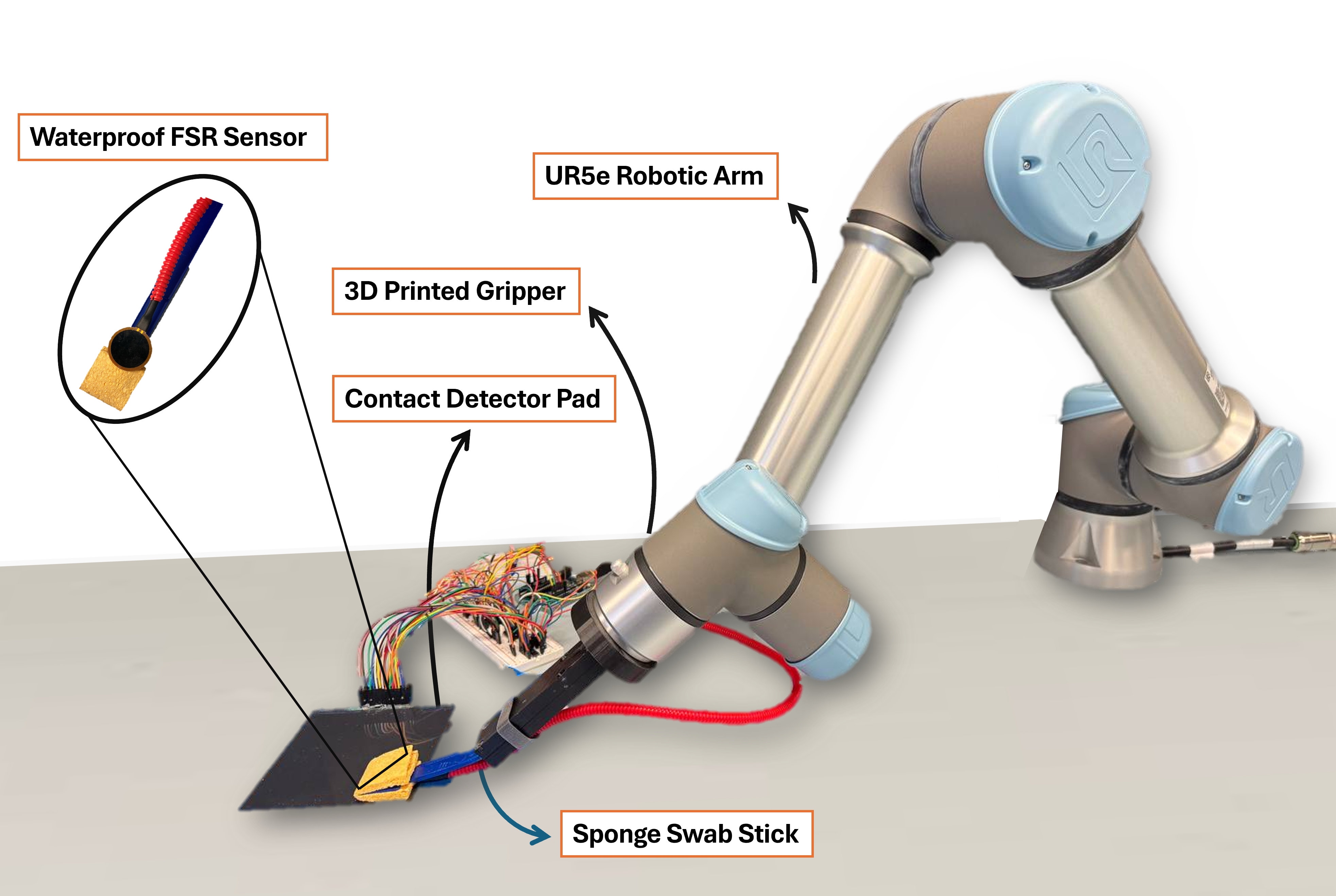} \caption{The robotic system setup consists of two different contact sensors (Waterproof FSR sensor and Contact Detector Pad) and a 3D printed swab holder gripper.} 
\label{fig:framework}
\end{figure}

The proposed system, illustrated in Fig. \ref{fig:framework}, consists of a UR5e robotic arm, a 3D-printed swab gripper, a standard food safety sponge swab (3M, Maplewood, MN, USA), and force sensitive resistor (fsr) sensors. a single-cell fsr and a matrix fsr pad (ShuntMode MatrixArray, Sensitronics, Skagit County, WA, US) were tested to assess their suitability for force-controlled swabbing.

To ensure optimal surface swabbing based on food safety standards, where precise force application and tool alignment are required, the robotic system needs to maintain consistent contact force throughout the swabbing path. The control system in this study can be divided into two key phases:
\begin{enumerate}
    \item \textbf{Contact Angle Alignment}: The robot adjusts the bending of the elastic swab stick handle until the embedded waterproof single-cell FSR sensor reaches the desired force threshold for effective microbial collection.
    \begin{figure}[h]
        \centering
        \includegraphics[width=1\linewidth]{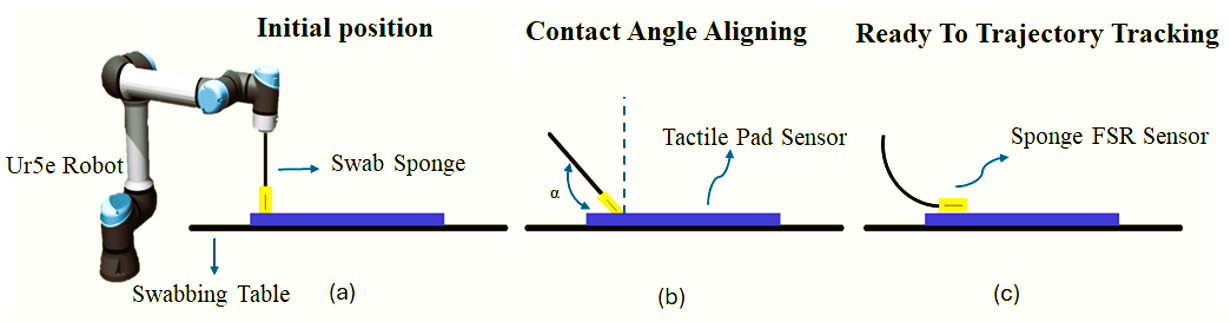}
        \caption{Contact angle alignment task: (a) step 1: initial position of trajectory tracking; (b) step 2: Angle alignment based on embedded FSR sensor feedback; (c) step 3: Ready to move to the next swabbing location following the predefined zigzag trajectory }
        \label{fig:enter-label}
    \end{figure}
    \item \textbf{Trajectory Tracking with Deformation Compensation}: The robot executes a zigzag swabbing pattern over a $10cm*10cm$ area per food safety standard, ensuring uniform force application while mitigating swab handle distortion and sponge pivoting during lateral movements.
\end{enumerate}
This framework enables reliable and repeatable robotic swabbing, maintaining both proper force distribution and trajectory precision throughout the process.

\subsection{Tactile Sensors Utilization}
\begin{figure}[h]
    \centering
    \includegraphics[width=1\linewidth]{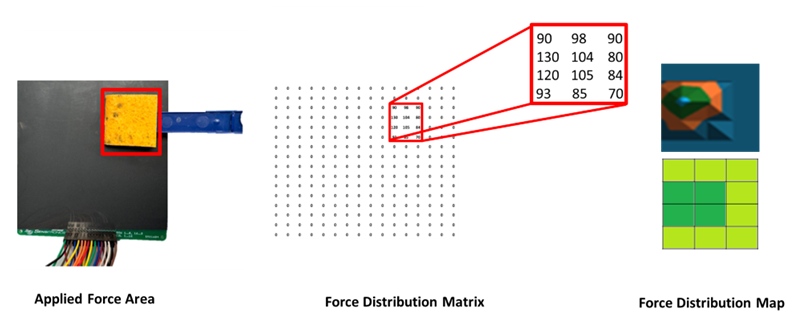}
    \caption{FSR Array Matrix Pad Contact Distribution Map results based on the sponge coverage contact}
    \label{fig:pad}
\end{figure}

The FSR pad, consisting of 256 sensing cells, was utilized to cover a $10cm*10cm$ active area, aligning with the standard swabbing zone. This sensor pad was initially employed to analyze human swabbing performance, generating a contact distribution map (Fig. \ref{fig:pad}) that highlighted inconsistencies in coverage and force application, even among experienced users. In trials involving over 20 trained participants, results indicated that human swabbing frequently left gaps in surface coverage and exhibited force inconsistencies, primarily due to the challenge of maintaining steady pressure throughout the process.

Although this FSR pad is effective for contact detection and force distribution monitoring, its repeatability is limited to around 20\%, making it unsuitable for precise force control \cite{flores2015resolution}. While ideal for contact-based force control, conventional 6-axis force/torque (F/T) sensors on the wrist suffer from significant drift when used with deformable tools such as the swab stick. To address these challenges in this study, a waterproof, embedded single-cell FSR sensor was integrated inside the sponge swab. This sensor provides direct contact force intensity feedback, offering a cost-effective, drift-resistant, and repeatable solution for force-controlled swabbing while overcoming the limitations observed in previous sensors.

\subsection{Framework Description}
\begin{figure}[h]
    \centering
    \includegraphics[width=1\linewidth]{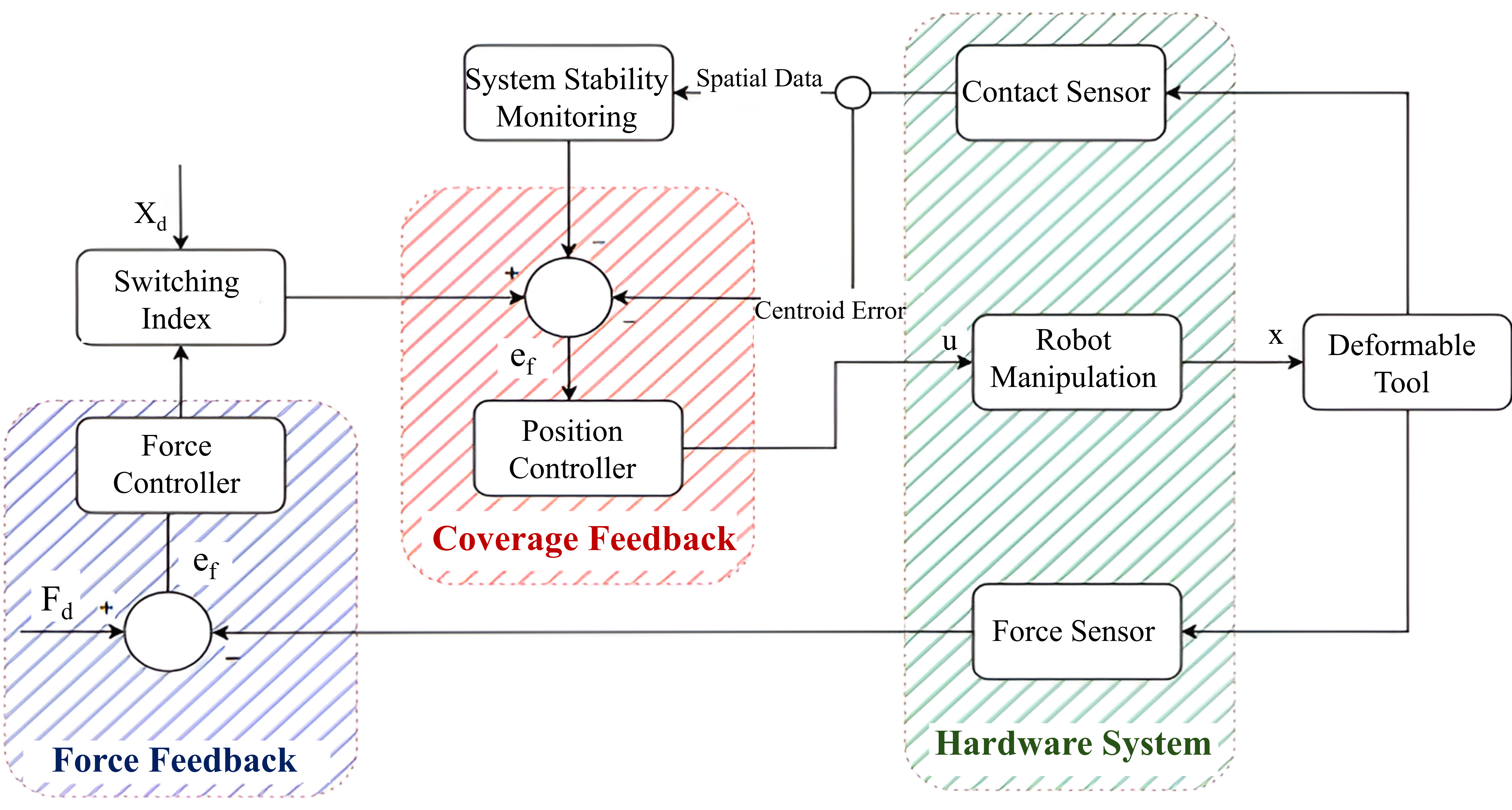}
    \caption{Control scheme of the proposed overall framework }
    \label{fig:flowchart}
\end{figure}

The proposed control framework is shown in Fig \ref{fig:flowchart}, which ensures precise control of the deformable swabbing tool by integrating force regulation and trajectory tracking through a structured control architecture. The system follows a dual-loop control strategy, where an inner loop focuses on force regulation, ensuring consistent contact pressure, while an outer loop governs trajectory tracking to maintain uniform coverage based on the position control. A switching mechanism dynamically selects between force and coverage control modes based on task requirements, allowing seamless adaptation to different stages of the swabbing process.

At the core of the control framework, the force controller regulates contact pressure by adjusting the robot’s rolling angle. This ensures that the swab maintains the desired force level despite the non-linearity from the deformable handle and sponge compression. This is particularly crucial for achieving effective microbial collection, as improper force distribution can lead to inconsistent swabbing results \cite{ismail2013methods}. In contrast, the coverage controller ensures that the swabbing motion follows the predefined trajectory while minimizing excessive deformation or unintended pivoting of the sponge, which could otherwise compromise surface coverage.

The switching mechanism enables real-time mode selection based on sensor feedback. When the system prioritizes force control, the trajectory tracking loop is disabled, allowing the robot to focus on maintaining a stable force profile. Conversely, when position control is required, the force controller is deactivated, enabling the system to execute smooth and precise motion along the designated path. This hierarchical approach prevents conflicts between force and position objectives, ensuring that both are met efficiently in different phases of the swabbing process.

This control architecture effectively balances compliance and precision, addressing the challenges associated with DTM. By integrating force and position control within a unified framework, the system ensures that both contact stability and tool deformation are maintained, making it well-suited for automated swabbing tasks where consistency and precision are critical.

\section{State Adaptive Optimal Angle Alignment and Contact Distribution Control}
Ensuring a consistent and optimal contact force between the deformable tool and the table is critical in achieving effective contact-aware control, a factor that differentiates robotic from human performance. The deformable nature of the sponge and the elastic handle introduces additional complexities, as their physical properties cause nonlinear interactions between force application and tool deformation. Furthermore, the inherent nonlinearity of the tactile sensor response exacerbates the challenge of force control, making conventional linear control strategies insufficient.

To address this issue, Koopman operator theory is employed to transform the inherently nonlinear dynamics of the tool-sensor system into a linear representation \cite{bruder2019nonlinear}. By lifting the system states into a higher-dimensional space, the Koopman approach enables the use of linear control techniques on an otherwise complex, nonlinear system. This transformation allows for a more structured and predictable model, improving the stability and accuracy of force regulation \cite{mauroy2016linear}.

Once the system is linearized through the Koopman operator, an optimal control scheme based on the LQR is designed to achieve precise force regulation. The LQR framework optimizes control inputs by minimizing a cost function that balances force error and control effort, ensuring smooth, energy-efficient, and stable adjustments of the robot’s states. This approach allows the robotic system to continuously adjust its orientation to maintain the target force, despite variations in sponge deformation and handle elasticity.

\subsection{ Operator-Based System Linearization}
To describe the system dynamics, the state vector is defined as 
$x_k = (\theta_k, p_k)^\top$, where $\theta_k$ represents the end-effector y-axis rolling angle, and $p_k = (x_k^p, y_k^p, z_k^p)^\top$ denotes the Cartesian position of the end-effector. The control inputs are six joint torques, represented as $u_k = (\tau_1, \tau_2, \dots, \tau_6)^\top$, and the system output is the force measurement from the embedded FSR sensor, denoted as $y_k$. The overall system follows the nonlinear discrete-time formulation:
\begin{equation}
    x_{k+1} = F(x_k, u_k)
\end{equation}
\begin{equation}
    y_k = g(x_k, u_k) + w_k
\end{equation}
where \( F \) represents the unknown nonlinear system dynamics, \( g \) describes the force response, and \( w_k \) accounts for measurement noise. 
The control challenge involves maintaining the correct rolling angle \( \theta_k \) to regulate force while ensuring trajectory adherence without excessive deformation of the swab handle.

The Koopman operator, denoted as \( K \), is an infinite-dimensional linear operator that advances observable functions \( \Psi(s) \) instead of evolving the system state directly:

\begin{equation}
    K \Psi(s) = \Psi(F(s))
\end{equation}

where \( s \in \mathbb{R}^N \) is the system state, \( F: \mathbb{R}^N \to \mathbb{R}^N \) represents the nonlinear state evolution, and \( \Psi(s) \) is the lifting function:

\begin{equation}
    \begin{aligned}
        \Psi(s) &= \begin{bmatrix} \psi_1(s), \psi_2(s), \dots, \psi_M(s) \end{bmatrix}^T, \\
       \Psi &: \mathbb{R}^N \to \mathbb{R}^M, \quad M > N.
    \end{aligned}
\end{equation}

Thus, rather than directly approximating \( F(s) \), the system evolves as:

\begin{equation}
    \Psi(s_{k+1}) = K_d \Psi(s_k)
\end{equation}

where \( K_d \) is the discrete-time Koopman operator, providing a globally valid linearization. The Koopman representation reformulates the nonlinear state evolution as:

\begin{equation}
    \Psi(x_{k+1}, u_{k+1}) \approx K_d \Psi(x_k, u_k)
\end{equation}

where \( \Psi(x_k, u_k) \) is a lifting function that maps the original states into a higher-dimensional space, capturing the complex interactions between the tool and the surface. In this work, we employ a combination of polynomial basis functions (degree 2) and radial basis functions (RBFs) to construct the lifting function:

\begin{equation}
    \Psi(x, u) =
    \begin{bmatrix}
        x \\
        u \\
        x^2 \\
        u^2 \\
        xu \\
        e^{-\|x - c\|^2}
    \end{bmatrix}
    \label{eq:lift}
\end{equation}

where polynomial terms capture state-dependent nonlinearities, while RBF kernels handle trajectory-dependent force variations.
Using Extended Dynamic Mode Decomposition (EDMD) \cite{williams2015data}, the Koopman operator $K_d$ is computed as:
\begin{equation}
    K_d = G^{+}  A_{EDMD}
    \label{eq:koopman}
\end{equation}
with
\begin{equation}
    G = \frac{1}{N} \sum_{k=1}^{N} \Psi(x_k, u_k) \Psi(x_k,u_k)^\top,
\end{equation}

\begin{equation}
    A_{EDMD} = \frac{1}{N} \sum_{k=1}^{N} \Psi(x_k, u_k) \Psi(x_{k+1}, u_{k+1})^\top.
\end{equation}
where $G^{+}$ represents the pseudoinverse of $G$. This transformation allows the nonlinear system to be expressed in linear state-space form, facilitating optimal control design. This leads to the Koopman-based system representation: 
\begin{equation}
    \begin{bmatrix} x_{k+1} \\ u_{k+1} \end{bmatrix} =
    \begin{bmatrix} A & 0 \\ B & I \end{bmatrix}
    \begin{bmatrix} x_k \\ u_k \end{bmatrix}
    \label{8}
\end{equation}
where $ A = K^T \frac{\partial \Psi}{\partial x}$ represents the state transition matrix, and $B = K^T \frac{\partial \Psi}{\partial u}$ represents the input influence matrix.

\subsection{SA-KLQR Controller}
\begin{figure}[h]
    \centering
    \includegraphics[width=1\linewidth]{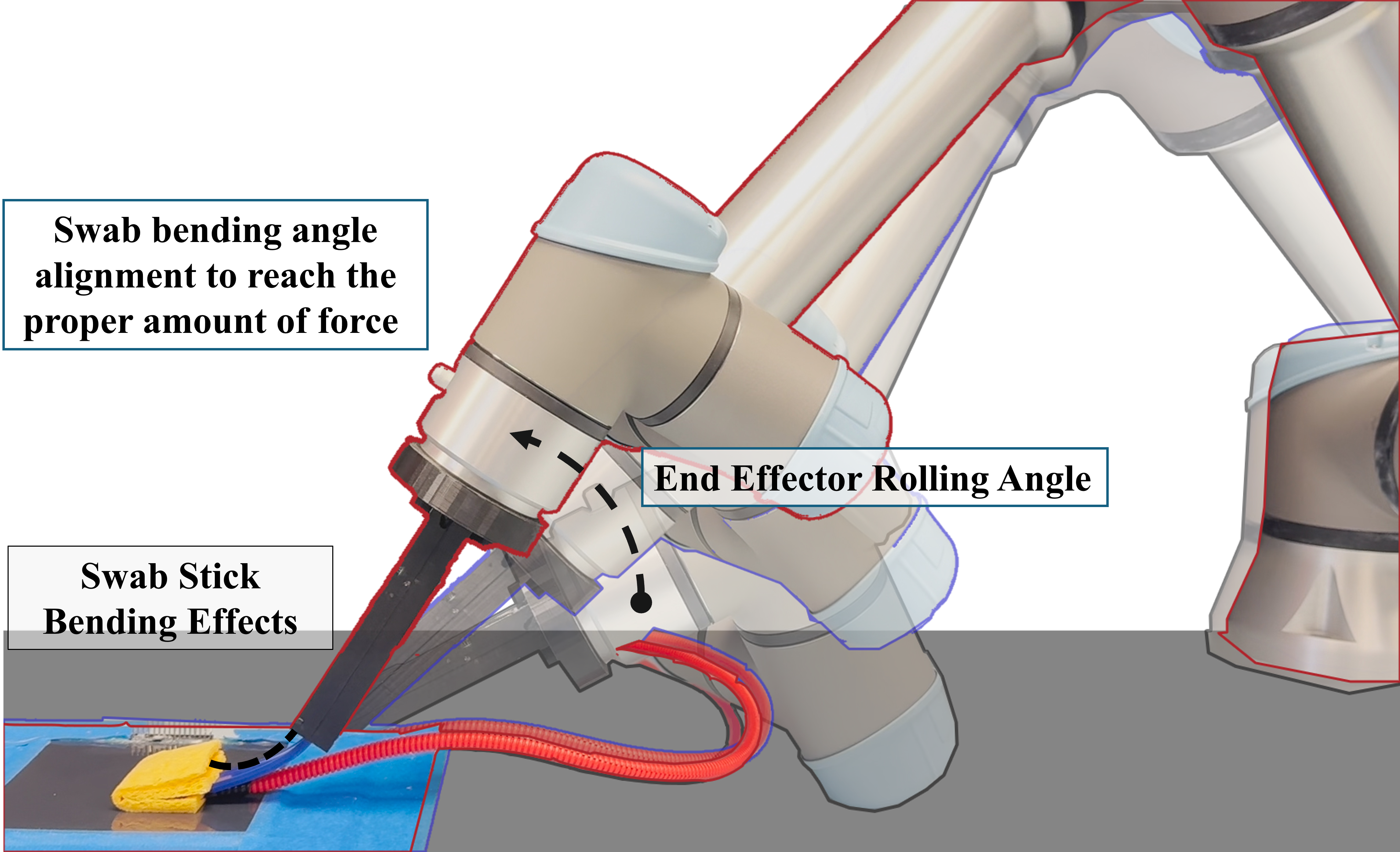}
    \caption{End-effector rolling angle and Koopman-based force regulation for maintaining stable contact pressure at the given location.}
    \label{fig:rolling}
\end{figure}

Once the Koopman representation is established, we develop the SA-KLQR controller for optimal force regulation. The transformed system follows the linearized dynamics:
\begin{equation}
    x_{k+1} = A x_k + B u_k
\end{equation}
\begin{equation}
    y_k = C x_k
\end{equation}
Where $A$ and $B$ can be provided by equation \ref{8} and $C$ maps the lifted state back to the sensor output. The optimal control law is obtained via LQR which minimizes a quadratic cost function \cite{shi2022deep}:
\begin{equation}
    J = \sum_{k=0}^{\infty} \left( x_k^T Q x_k + u_k^T R u_k \right)
    \label{eq:J}
\end{equation}
where $Q\succ 0$ penalizes force deviation from the reference trajectory, and $ R \succ 0$ penalizes excessive control effort. The optimal LQR control law takes the state-feedback form:
\begin{equation}
    u_k = -K_k x_k + K_r v + K_I \int_{0}^{t} e_{\tau} d\tau
    \label{eq:u_k}
\end{equation}
where \( K_k \) and \( K_r \) are optimal feedback gains derived from solving the discrete-time Riccati equation:
\begin{equation}
    P = Q + A^\top P A - A^\top P B (R + B^\top P B)^{-1} B^\top P A
\end{equation}
\( K_k \) is the integral gain to compensate for steady-state errors \cite{athans2007optimal}.
Fig. \ref{fig:rolling} illustrates the end-effector’s rolling angle alignment, regulated by real-time FSR feedback for precise force control. The rotation follows a quaternion-based model, ensuring smooth adjustments without singularities. The incremental roll-axis rotation is given by:
\begin{equation}
    q_{\text{increment}} = \left( \sin\left(\frac{\theta}{2}\right), 0, 0, \cos\left(\frac{\theta}{2}\right) \right)
\end{equation}
with the updated orientation computed as:
\begin{equation}
    q_{\text{new}} = q_{\text{current}} \otimes q_{\text{increment}}
    \label{eq:q}
\end{equation}
where \( \otimes \) denotes quaternion multiplication. This ensures stable force regulation within the Koopman-based control framework.

Since force dynamics, sensor response, and arm inertia vary across the swabbing trajectory, a single Koopman operator cannot fully capture these variations. Instead of continuous interpolation, which is computationally demanding, we implement a state-dependent Koopman switching mechanism. This approach selects the most relevant Koopman operator in real time based on the robot’s current state, ensuring efficient and accurate force regulation.

\begin{figure}
    \centering
    \includegraphics[width=0.85\linewidth, height=6cm]{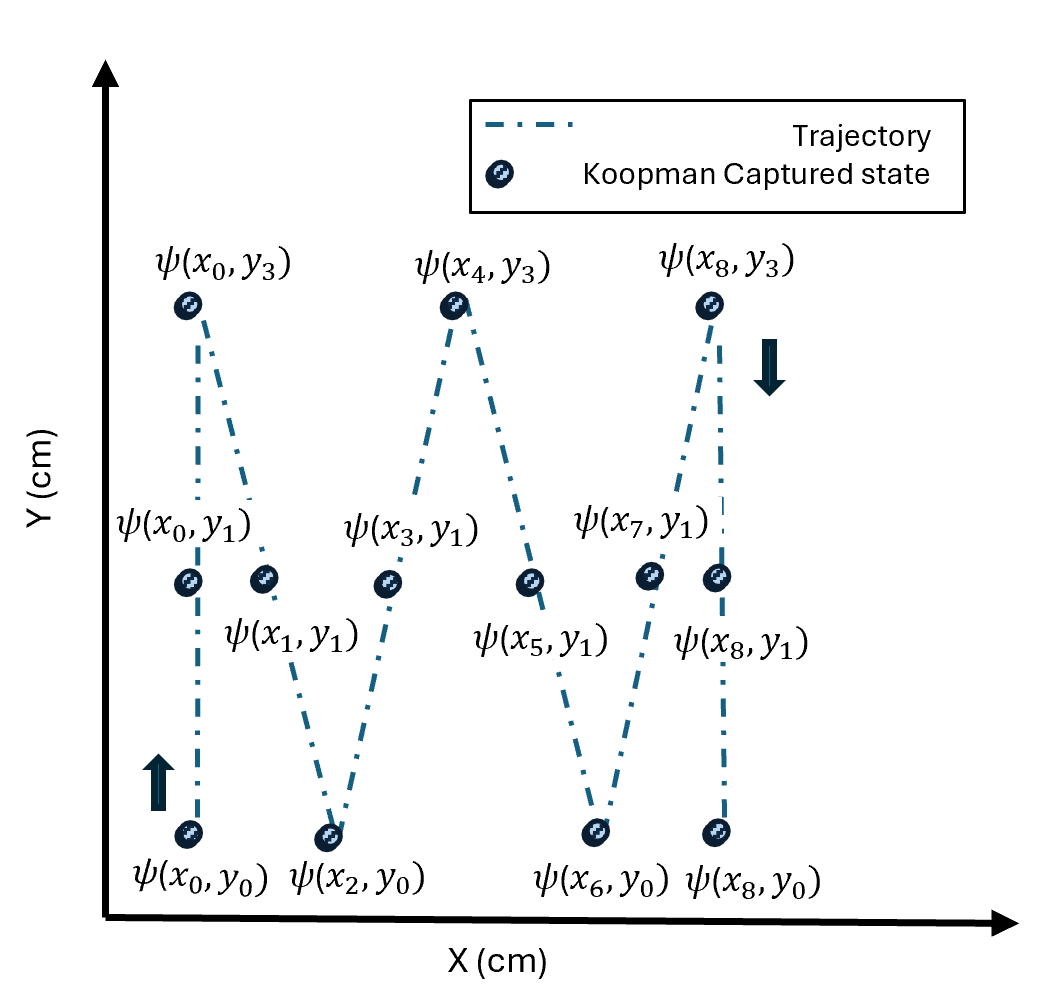}
    \caption{Region-Based Koopman Segmentation Along the Zigzag Trajectory: Each region has a localized Koopman operator $k_i$ to adaptively capture varying force dynamics and tool compliance for smooth swabbing control.}
    \label{fig:trajectory}
\end{figure}

The state space is divided into multiple operating regions, each associated with a precomputed Koopman operator $k_i$. These regions correspond to distinct force regimes, wrist angles, or surface interactions where system dynamics exhibit significant variation. These Koopman operators are derived offline using EDMD, ensuring that each operator accurately represents local system behavior as shown in Fig \ref{fig:trajectory}. At runtime, the robot identifies the nearest operating region based on its current state \( s_k \) and selects the corresponding Koopman operator \( K_{\text{selected}} \), given by:

\begin{equation}
    K_{\text{selected}} = \arg \min_{i} \| x_k - c_i \|
    \label{eq:k1}
\end{equation}
where \( c_i \) represents the center of region \( i \).
To prevent abrupt control jumps when switching between Koopman operators, we employ a buffered transition function, ensuring gradual adaptation:

\begin{equation}
    K_{\text{current}} = (1 - \beta) K_{\text{previous}} + \beta K_{\text{selected}}
    \label{eq:k2}
\end{equation}

where:  \( K_{\text{previous}} \) is the Koopman operator from the previous time step. \( K_{\text{selected}} \) is the Koopman operator for the new region. \( \beta \) is an adaptive smoothing factor that determines how quickly the transition occurs.  Algorithm \ref{alg:SA-KLQR} summarizes the proposed linear model control construction process.

\begin{algorithm}
\caption{\textbf{SA-KLQR controller system.}}
\label{alg:SA-KLQR}
\begin{algorithmic}
    \STATE \textbf{Input:} \(x_k, u_k, y_k\)  for \( s = 1, \dots, S \) and \(K_i\) for  \( i = 1, \dots, I \). 
    \STATE \textbf{Step 1:} Lift data via (\ref{eq:lift}).
    \STATE \textbf{Step 2:} Compute the Koopman operator using via (\ref{eq:koopman}).
    \STATE \textbf{Step 3:} Define the LQR cost function via (\ref{eq:J}) and compute the optimal control input via (\ref{eq:u_k}) ensuring stable force control.
    \STATE \textbf{Step 4:} Identify the nearest Koopman operator based on the current state by \ref{eq:k1}.
    \STATE \textbf{Step 5:} Ensure smooth transitions between Koopman operators using \ref{eq:k2}.
    \STATE \textbf{Step 6:} Apply quaternion-based rolling angle correction for stable orientation control (\ref{eq:q}).
    \STATE \textbf{Output:} Optimal control input $u_k$ for precise force regulation
\end{algorithmic}
\end{algorithm}

\subsection{Centroid-Based Fuzzy Force Regulation}
To achieve effective force regulation, the proposed controller maintains a rigid tool model for rolling angle control. However, due to the tool’s elasticity, abrupt direction changes or sideward movements cause unintended distortions in force distribution, shifting the force centroid away from the ideal location. This centroid drift accumulates over time, propagating errors into the force-tracking loop and leading to excessive deformation or tool instability.

Although gripper design modifications can mitigate this issue to some extent by improving mechanical stability, a purely hardware-based approach is insufficient for ensuring consistent and controlled force application. Instead, the proposed framework enables real-time monitoring and regulation of force distribution across the in contacted tool surface, actively compensating for tool compliance. By leveraging centroid-based force tracking and fuzzy entropy analysis, the system dynamically corrects tool pivoting and deformation, ensuring stability throughout the swabbing process. This integration of data-driven force monitoring and adaptive control enhances the reliability of robotic swabbing, minimizing force inconsistencies and maximizing sample collection efficiency.

The force distribution across the FSR sensor pad, shown in Fig \ref{fig:pad}, is represented as a discrete matrix of force values \( f_{ij} \), where \( (x_i^c, y_j^c) \) denote the spatial coordinates of force application. The centroid of the force distribution is computed as:

\begin{equation} C_x = \frac{\sum_{i=1}^{n} \sum_{j=1}^{m} x_k^c f_{ij}}{\sum_{i=1}^{n} \sum_{j=1}^{m} f_{ij}}, \quad C_y = \frac{\sum_{i=1}^{n} \sum_{j=1}^{m} y_j^c f_{ij}}{\sum_{i=1}^{n} \sum_{j=1}^{m} f_{ij}} \label{eq:C} \end{equation}

where \( C_x \) and \( C_y \) denote the centroid coordinates in the \( x \) and \( y \) directions, respectively. Here,\( n \) and \( m \) define the number of force-sensing cells along the \( x-axis \) and \( y-axis \), ensuring consistency with the spatial definition of the sensing pad. The denominator normalizes the force-weighted summation to ensure that the centroid location is accurately represented. 

The centroid error \( D \)  is defined as the Euclidean distance between the computed centroid \( (C_x, C_y) \) and the desired force application center \( (C_x^*, C_y^*) \), given by:

\begin{equation}
    D = \sqrt{(C_x - C_x^*)^2 + (C_y - C_y^*)^2}
    \label{eq:D}
\end{equation}
where \( C_x^* \) and \( C_y^* \) represent the ideal centroid coordinates. A lower centroid error indicates a more balanced force application, while higher values suggest instability and tool misalignment.

A lower centroid error indicates a more balanced force application, whereas higher values suggest instability and tool misalignment. To quantify the temporal consistency of centroid error, we employ Fuzzy Entropy (FuzzyEn) \cite{al2001fuzzy}, which provides a measure of complexity and unpredictability in the centroid error time series. Fuzzy entropy is computed as:

\begin{equation}
    \text{FuzzyEn}(m, r, N) = - \lim_{N \to \infty} 
    \ln\left(\frac{\sum_{i=1}^{N-m} e^{-\| D_i - D_j \| / r}}
    {N-m} \right)
    \label{eq:fuzzyEn}
\end{equation}

where \( m \) is the embedding dimension, controlling the pattern length.
 \( r \) is the similarity threshold, defining tolerance for variation.
\( N \) is the time series length. A lower FuzzyEn value indicates a predictable and regular centroid error pattern, characteristic of balanced force application, while higher values signify increasing variability, highlighting unstable swabbing dynamics. Algorithm \ref{alg:centroid} formalizes this Centroid-Based Fuzzy Force Regulation  approach, detailing real-time monitoring, correction, and stability evaluation to ensure consistent force application throughout the swabbing trajectory

\begin{algorithm}
\caption{\textbf{Centroid-Based Fuzzy Force Regulation.}}
\label{alg:centroid}
\begin{algorithmic}
    \STATE \textbf{Input:} Force matrix $f_{ij}$ with spatial coordinates \(x_i, y_j\), Desired force centroid $(C_{center x}, C_{center y})$ and Embedding dimension $m$ similarity threshold $r$, time series length $N$.
    \STATE \textbf{Step 1:} Calculate the force distribution centroid using (\ref{eq:C}). following by Measure centroid deviation from the ideal reference via (\ref{eq:D}).
    \STATE \textbf{Step 2:} Compute Fuzzy Entropy to analyze temporal consistency via \ref{eq:fuzzyEn}
    \STATE \textbf{Step 3:} \begin{itemize}
    \item \textbf{If} \( D \) exceeds a predefined threshold:
    \begin{itemize}
        \item Adjust rolling angle to correct tool alignment via Algorithm 1.
        \item Modulate force application to ensure uniform pressure.
    \end{itemize}
    
    \item \textbf{If} FuzzyEn is high (unstable force application):
    \begin{itemize}
        \item Increase correction frequency.
        \item Modify trajectory for smoother force transitions.
    \end{itemize}
\end{itemize}

    \STATE \textbf{Step 4:} Continuously update centroid position and force error and Maintain real-time monitoring for adaptive correction.
    \STATE \textbf{Output:} Optimized force distribution with minimized centroid error.
\end{algorithmic}
\end{algorithm}

\section{EXPERIMENTS, RESULTS AND DISCUSSIONS}
\subsection{Embedded FSR Sensor Behavior Evaluation}
\begin{figure}[h]
    \centering
    \includegraphics[width=1\linewidth]{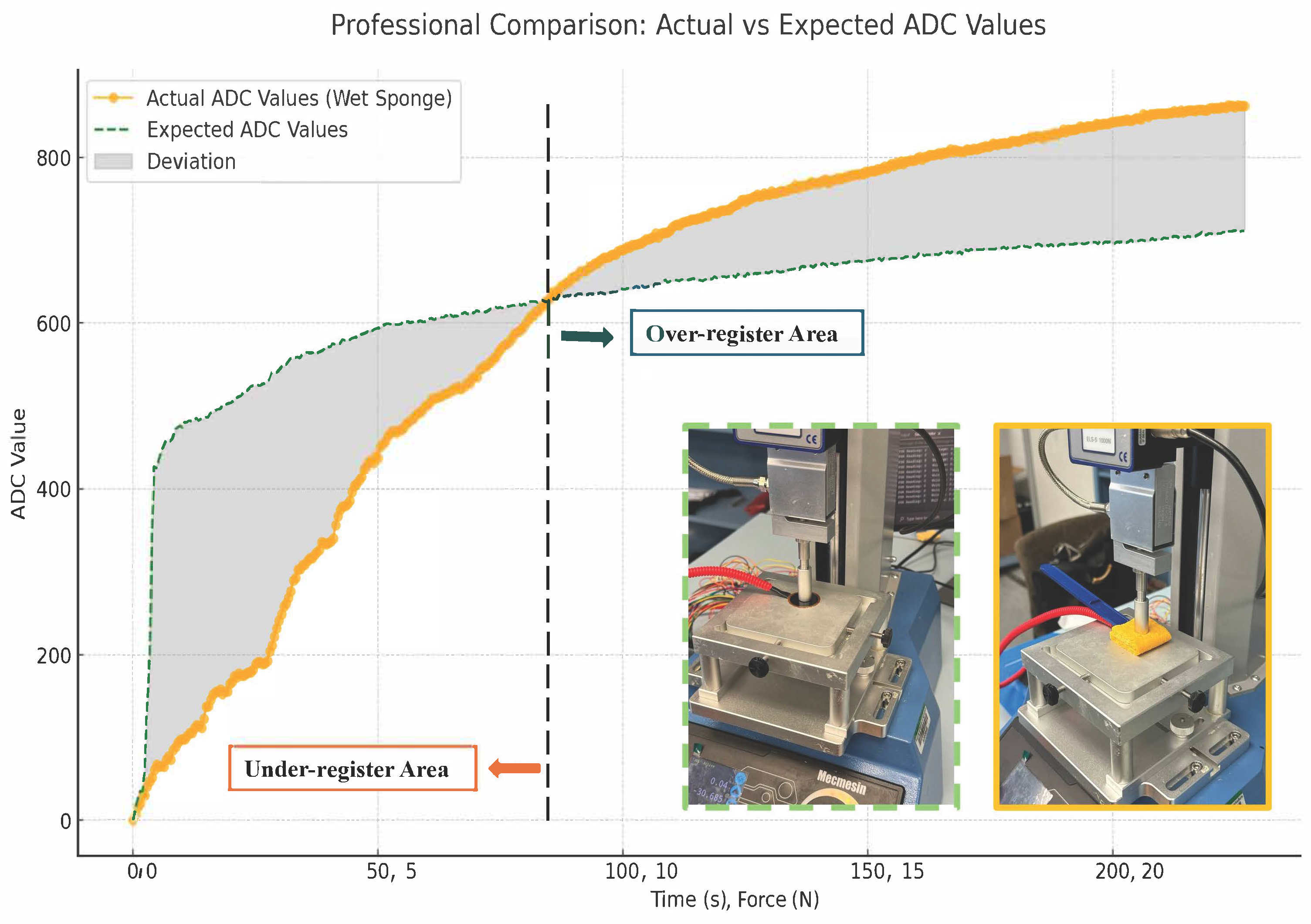}
    \caption{Comparison of actual and expected ADC values for the FSR sensor with a deformable sponge. The under-registration phase at low forces results from the sponge absorbing force, while over-registration at high forces occurs due to increased compression efficiency. The images depict the experimental setup for direct sensor exposure (left) and embedded sponge testing (right).}
    \label{fig:compare}
\end{figure}

To evaluate the performance of the force sensor under variable compression by a deformable medium, a series of tests were conducted using an advanced texture analyzer (TMS-Pro, Food Technology Corporation (FTC), VA, USA). A controlled force ranging from 0 to 25 N was applied over a duration of 250 seconds, and the sensor’s response was recorded under two conditions: directly exposed and embedded within a sponge. The results, shown in Fig. \ref{fig:compare}, illustrate the nonlinear effects of the sponge on force transmission.

The data reveal two distinct response phases due to the sponge’s compressibility. In the initial phase, the sensor under-registers force as the sponge absorbs part of the applied load, leading to a dampened measurement, particularly at lower force levels where the sponge remains partially compressed. As force increases, the response curve enters an over-registration phase, where the sponge, once nearing full compression, enhances force transmission efficiency, resulting in higher ADC values than expected for equivalent direct force applications.

Fig. \ref{fig:compare} compares the actual and expected ADC values for the embedded sensor, highlighting deviations caused by the sponge. The under-registration phase is most prominent at lower forces, while over-registration occurs at higher forces due to increased stiffness as the sponge reaches full compression. This characterization is critical for real-time force compensation, ensuring consistent and accurate force regulation in robotic swabbing applications.
\subsection{Koopman Operator Performance Evaluation}
To validate the effectiveness of the Koopman-based system linearization, multiple experiments were conducted to analyze its ability to model the nonlinear force dynamics of the robotic swabbing process. The accuracy of the Koopman-based system representation depends on the choice of observables. While deep neural networks have been explored for automatic discovery of Koopman function observables \cite{abraham2019active}, their high computational cost and lack of interpretability pose challenges for real-time robotic control applications. Instead, we employ a combination of polynomial basis functions and radial basis functions (RBFs), which provide a structured, computationally efficient, and physically interpretable representation of system dynamics. Polynomials capture global variations in force dynamics, while RBF kernels enable localized adaptability, ensuring robust force control in the presence of tool compliance.
table \ref{tab:methods_comparison} shows the tested observable method on prediction of different states.

\begin{table*}[ht]
  \centering
  \caption{Comparison of Different Observable Methods}
  \label{tab:methods_comparison}
  \begin{tabular}{@{}lcccccp{2cm}@{}}
    \toprule
    Observable Method  & Avg. RMSE (N) & Avg. $R^2$ Score & Avg. MAE (N) & Remarks \\ 
    \midrule
    Polynomial (Degree 2) & 0.06051 & 0.789 & 0.06643 &  Performs well but struggles with complex force variations. \\
    Polynomial (Degree 3)  & 0.04255 & 0.762 & 0.00465 &  More flexible but introduces slight overfitting. \\
    Radial Basis Functions (RBF)  & 0.01589 & 0.812 & 0.00331 &  Best performance overall, smooth transition handling. \\
    Fourier Basis  & 0.08202 & 0.780 & 0.08539 &  Captures periodic  but less effective for abrupt force changes. \\
    Gaussian Mixture Models (GMM) \cite{cao2024distributionally} & 0.01997 & 0.795 & 0.00534 & Good for force variations but computationally intensive. \\
    SINDy-based Koopman \cite{brunton2016discovering} & 0.00991 & 0.804 & 0.00349 &  Limited Handling of High-Dimensional Systems. \\
    Proposed combined method & 0.00608 & 0.960 & 0.00243  & captures global trends, enhances local adaptability.  \\ \bottomrule
  \end{tabular}
  \label{tab:1}
\end{table*}

\begin{figure}[h]
    \centering
    \includegraphics[width=1\linewidth]{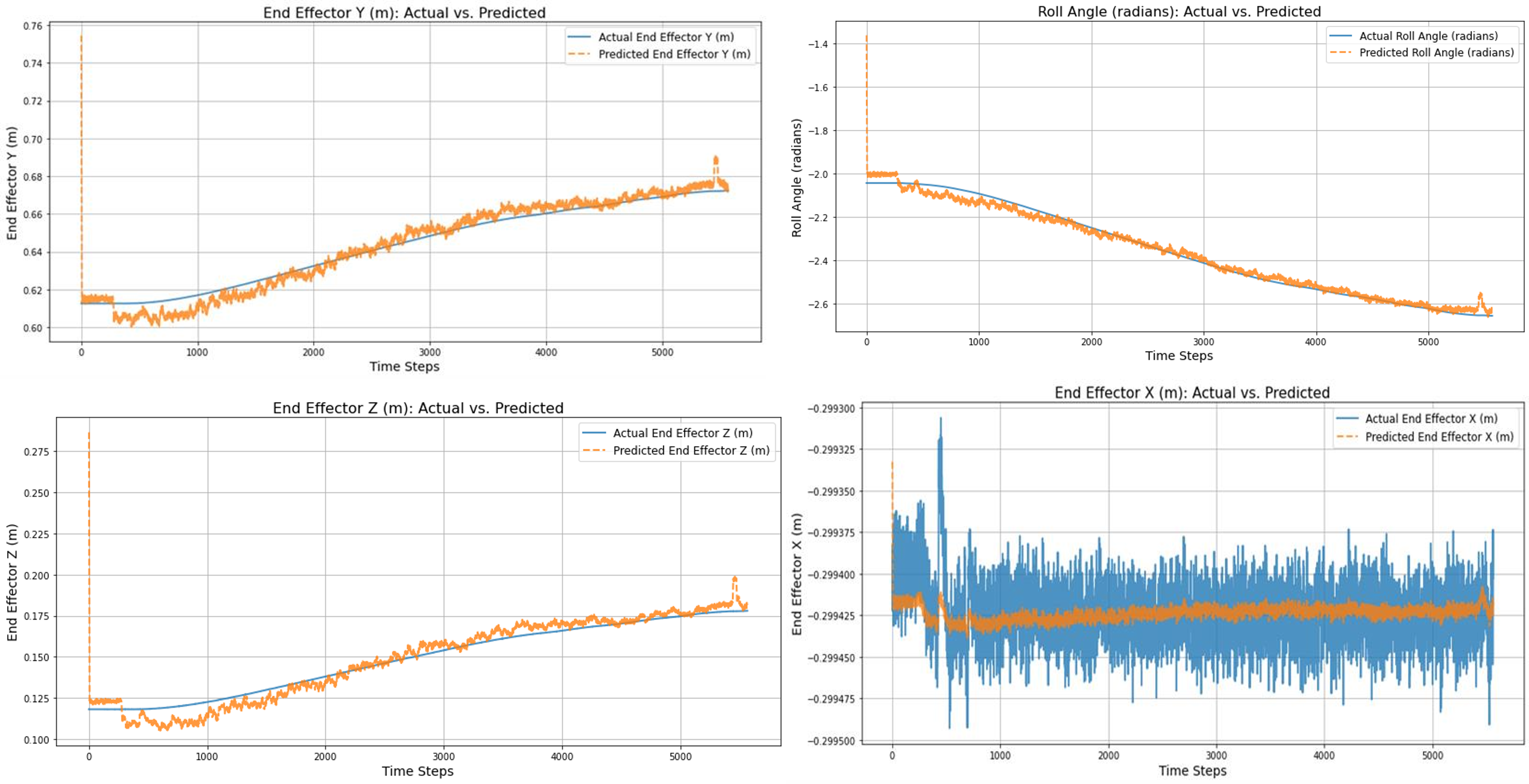}
    \caption{ Koopman-based prediction of end-effector states, including roll angle and Cartesian positions $(x_k^p, y_k^p, z_k^p)$. }
    \label{fig:states}
\end{figure}

\begin{figure}[h]
    \centering
    \includegraphics[width=1\linewidth]{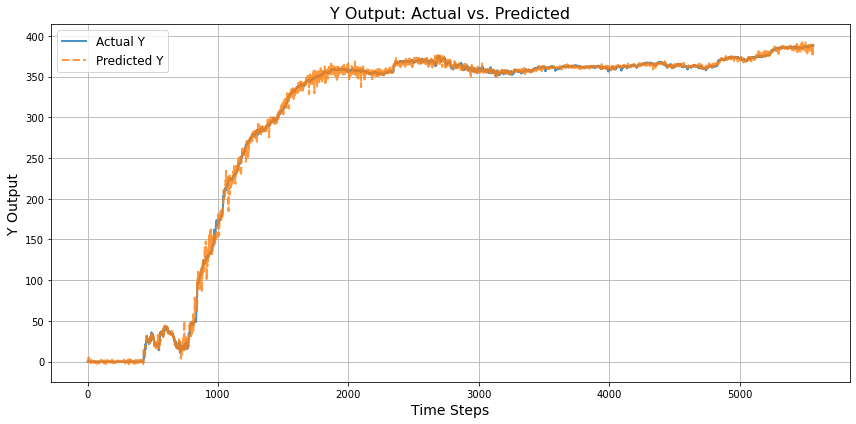}
    \caption{Prediction of sensor output in the wet sponge scenario.}
    \label{fig:sensor_tracking}
\end{figure}

Fig. \ref{fig:states} validates the Koopman-based model’s predictive accuracy across key robotic states, including roll angle and Cartesian positions $(x_k^p, y_k^p, z_k^p)$. The predicted roll angle closely follows the actual trajectory, demonstrating precise rotational modeling for force regulation. Similarly, the end-effector Y and Z predictions align well with actual values, effectively capturing vertical and lateral positional variations. While the X position exhibits fluctuations due to high-frequency noise, the model successfully predicts the mean behavior. These results confirm the Koopman model’s ability to accurately predict states and control inputs, providing a strong foundation for optimal control and force adaptation in robotic swabbing.
The Koopman-based model effectively tracks the sensor output in the wet sponge scenario, achieving an MSE of 0.0002, R² of 0.9590, and MAE of 0.0071. As shown in Fig. \ref{fig:sensor_tracking}, the predicted force values closely align with actual measurements, demonstrating the model’s ability to capture nonlinear sensor dynamics and force variations. The high R² score confirms strong predictive accuracy, making the model suitable for real-time force adaptation in robotic swabbing.

\subsection{SA-KLQR Force Control and Trajectory Tracking}
Extensive experiments were conducted to evaluate the performance and robustness of the proposed SA-KLQR force control scheme in regulating contact force during robotic swabbing. The control objective is to maintain the desired contact force $F_{desired}$ while ensuring stable trajectory execution. To benchmark performance, we compare SA-KLQR with Proportional-Integral-Derivative (PID) and SMC, two widely used controllers in force regulation.
The PID controller follows:
\begin{equation}
u_{PID} = K_P e + K_I \int_0^t e \, d\tau + K_D \frac{de}{dt}
\end{equation}
where \( e = F_{\text{desired}} - F_{\text{measured}} \), and \( K_P, K_I, K_D \) are the proportional, integral, and derivative gains, respectively. The PID controller gains \( K_P, K_I, K_D \) were initially determined using the Ziegler-Nichols method \cite{ziegler1942optimum} and further fine-tuned based on empirical observations to minimize overshoot, reduce settling time, and optimize steady-state accuracy.

The SMC (Sliding Mode Control) controller is formulated as:
\begin{equation}
u_{SMC} = \frac{1}{b_0} \left( \ddot{r} - f_0(x, \dot{x}) \right) + K_1 e + K_2 \dot{e} + \epsilon \, \text{sign}(s) + K_s s
\end{equation}
where \( s = e + \lambda_1 \dot{e} + \lambda_2 \int e \, dt \) is the sliding surface, and \( f_0(x, \dot{x}) \) represents nominal system dynamics.

Each controller was tested under identical conditions to assess force tracking accuracy, stability, and robustness across different trajectories. The experimental setup is illustrated in Fig. \ref{fig:framework}. The robot followed a sinusoidal force trajectory, ensuring that the force remains non-negative, defined as:
\begin{equation}
F_{\text{desired}}(t) = F_0 + F_{amp} \cdot \frac{\sin(2\pi \omega t) + 1}{2}
\end{equation}
where \( F_0 \) is the nominal contact force, \( F_{amp} = 10 \text{N}\) is the amplitude, and \( \omega \) is the frequency of oscillation. To fully evaluate the controllers under both steady and rapid force variations, two test cases were considered:

\begin{itemize}
    \item Low-frequency case: \( F_{amp} = 10 \, \text{N} \), \( \omega = 0.5 \, \text{Hz} \) (steady-state tracking).
    \item High-frequency case: \( F_{amp} = 10 \, \text{N} \), \( \omega = 2 \, \text{Hz} \) (rapid variation tracking).
\end{itemize}

Each controller was evaluated over two full cycles for both low- and high-frequency sinusoidal reference trajectories, recording Root Mean Squared Error (RMSE) and Mean Absolute Error (MAE) to assess steady-state accuracy and dynamic adaptability. An optimal controller should exhibit minimal tracking error, low phase lag in high-frequency conditions, and smooth, stable performance in low-frequency cases.

\begin{figure}
    \centering
    \includegraphics[width=1\linewidth]{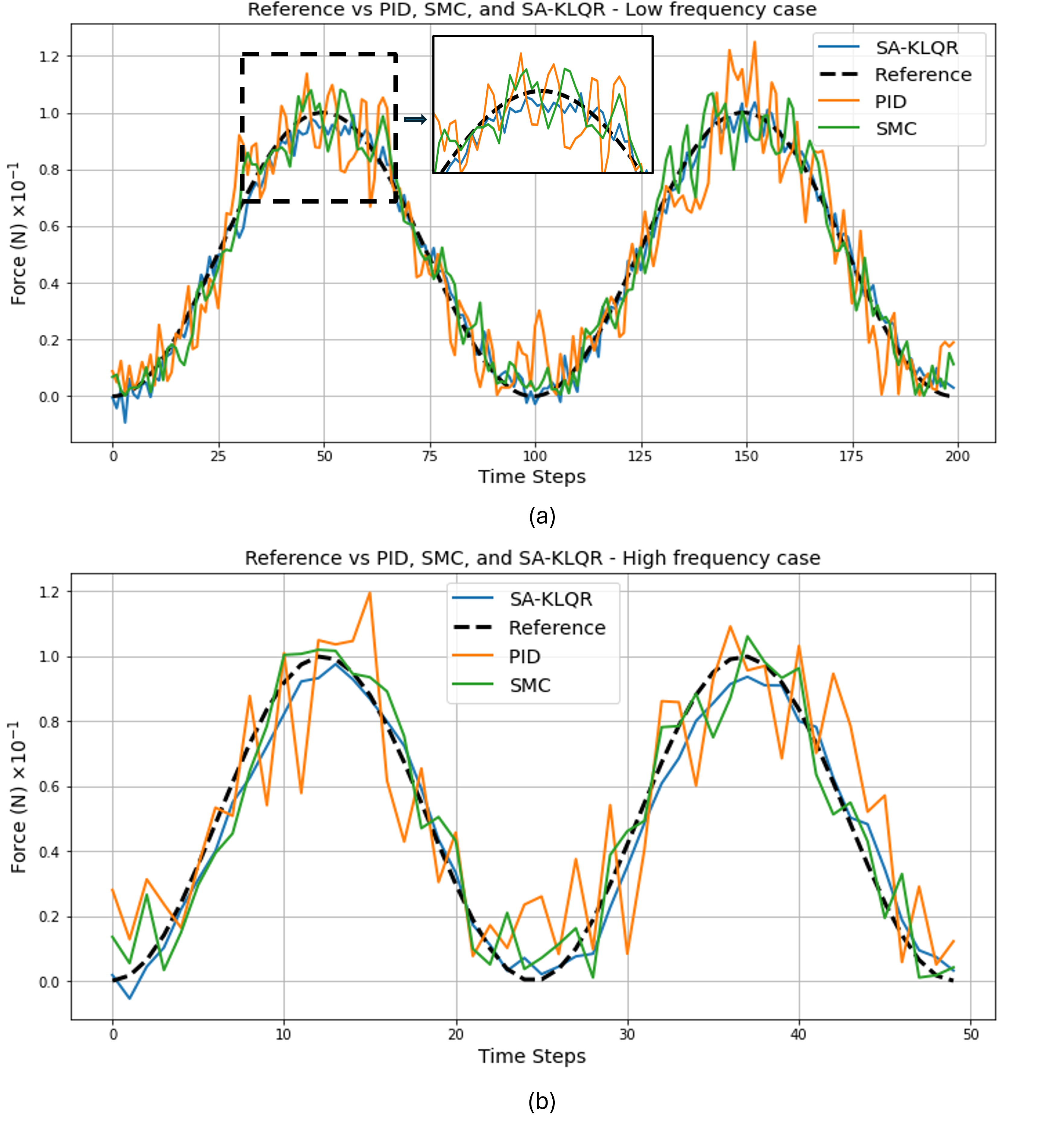}
    \caption{Force tracking performance comparison among SA-KLQR, PID, and SMC controllers for sinusoidal reference trajectories. (a) Low-frequency case: SA-KLQR maintains precise tracking with minimal error, while PID shows noticeable lag, and SMC introduces oscillatory behavior. (b) High-frequency case: SA-KLQR maintains accurate tracking, whereas PID struggles with phase lag, and SMC produces excessive oscillations. The inset in (a) highlights SA-KLQR's superior force regulation in finer details.}
    \label{fig:sins}
\end{figure}

\begin{figure}[h]
    \centering
    \includegraphics[width=1\linewidth]{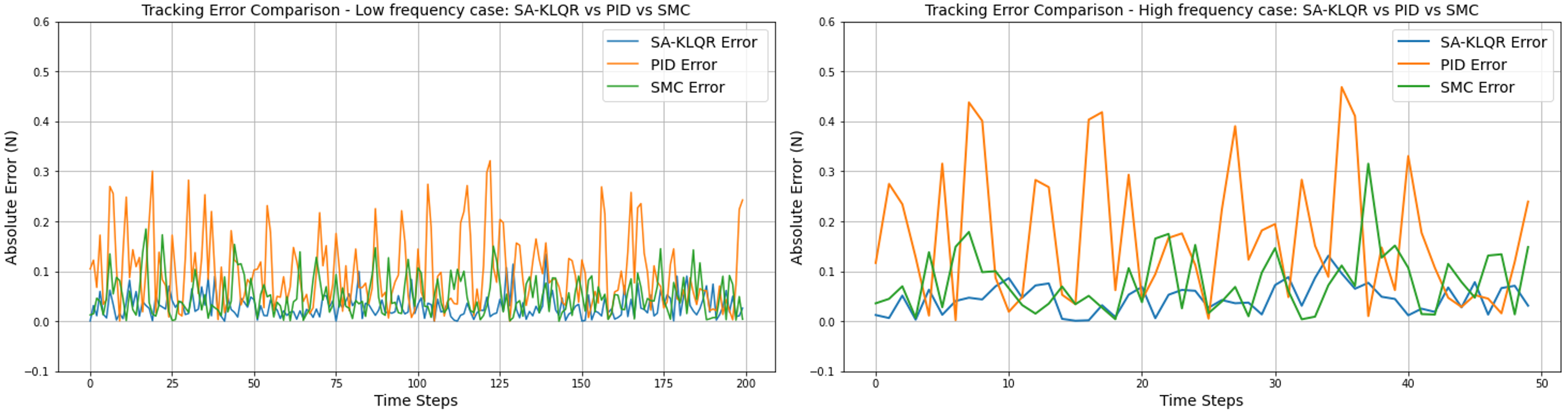}
    \caption{Absolute force tracking error comparison for low-frequency (left) and high-frequency (right) sinusoidal reference trajectories using SA-KLQR, PID, and SMC controllers. SA-KLQR consistently demonstrates lower error magnitudes and reduced fluctuations, particularly in high-frequency cases, showcasing its superior robustness and tracking accuracy.}
    \label{fig:sin-er}
\end{figure}

SA-KLQR consistently achieved the lowest RMSE and MAE, demonstrating superior force tracking, reduced phase lag, and enhanced stability compared to PID and SMC. Figure \ref{fig:sins} illustrates the force tracking performance, where SA-KLQR closely follows the desired force trajectory with minimal deviation. In low-frequency tracking, SA-KLQR provides precise force regulation with smooth transitions, while PID introduces noticeable lag, and SMC exhibits oscillations. In the high-frequency scenario, SA-KLQR maintains accurate tracking, whereas PID struggles with phase lag, and SMC generates excessive oscillations, confirming the benefits of the Koopman-based adaptive control strategy. 
Figure \ref{fig:sin-er} presents the absolute force tracking errors for both cases. SA-KLQR outperforms PID and SMC in both low- and high-frequency tracking. In low frequency, PID shows error spikes, and SMC has higher variance, while SA-KLQR maintains stability. In high frequency, PID struggles with oscillations, and SMC remains inconsistent, whereas SA-KLQR adapts effectively with minimal fluctuations, ensuring robust force regulation.

\begin{figure}[h]
    \centering
    \includegraphics[width=1\linewidth]{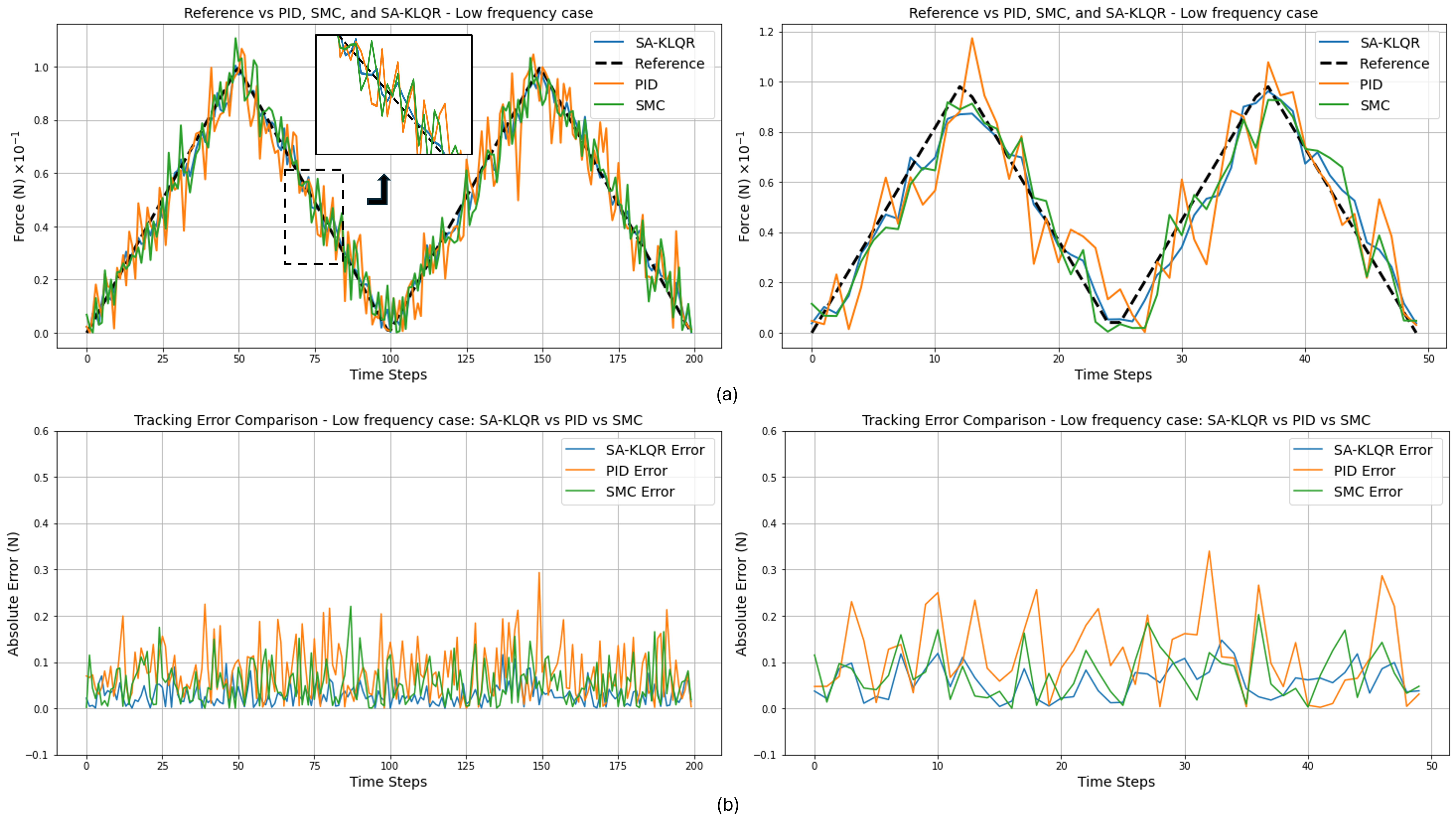}
    \caption{Force tracking performance and absolute error comparison for a low-frequency triangular wave trajectory using SA-KLQR, PID, and SMC controllers. (Top-left) Force tracking results for SA-KLQR, PID, and SMC controllers with an inset highlighting a transition region. (Top-right) Magnified view of the reference and tracking responses. (Bottom-left) Absolute force tracking error for the full trajectory. (Bottom-right) Zoomed-in view of the tracking error, showing that SA-KLQR maintains the lowest error, while PID suffers from lag and SMC introduces oscillatory behavior at transition points.}
    \label{fig:tri-all}
\end{figure}

The second set of experiments evaluated force tracking performance using a triangular wave trajectory, designed to assess each controller's ability to handle abrupt force transitions. Figure \ref{fig:tri-all} presents the tracking results and absolute error comparisons for both low- and high-frequency cases.

In the low-frequency case (top-left), SA-KLQR demonstrated precise tracking with smooth transitions, maintaining minimal deviation from the reference trajectory. The PID controller exhibited noticeable lag at sharp transitions, struggling to adapt to sudden force changes, while the SMC controller introduced oscillatory behavior, particularly near discontinuities. The absolute error analysis (bottom-left) further highlights SA-KLQR’s superior accuracy, with consistently lower error magnitudes compared to PID and SMC.
For the high-frequency case (top-right), tracking accuracy deteriorated across all controllers due to the increased difficulty in handling rapid force variations. The PID controller exhibited significant phase lag, while SMC reduced the error magnitude but still showed fluctuations at transition points. In contrast, SA-KLQR maintained the lowest error levels (bottom-right), effectively adapting to sharp force variations and minimizing oscillatory effects, reinforcing its robustness in dynamic force tracking tasks.

Table \ref{tab:error_comparison} presents the force tracking performance of SA-KLQR, PID, and SMC across sinusoidal and triangular trajectories at different frequencies. SA-KLQR outperforms both methods, reducing MaxAE and RMSE significantly. PID consistently exhibits errors approximately 55\% higher than SA-KLQR, struggling with tracking precision and force transitions. SMC performs better than PID but still maintains around 30\% higher errors than SA-KLQR. The performance gap is more pronounced in high-frequency cases, where PID suffers from increased phase lag and oscillations, while SA-KLQR maintains superior stability. These results further confirm the adaptability of SA-KLQR in dynamic force regulation tasks.

\begin{table}[h]
    \centering
    \caption{MaxAE and RMSE Comparison for Sine and Triangle Trajectories at Low and High Frequencies}
    \label{tab:error_comparison}
    \renewcommand{\arraystretch}{1.1}
    \setlength{\tabcolsep}{4pt} % Adjust column spacing
    \begin{tabular}{l|cc|cc}
        \hline
        \multirow{2}{*}{Controller} & \multicolumn{2}{c|}{Sine} & \multicolumn{2}{c}{Triangle} \\
        & MaxAE & RMSE & MaxAE & RMSE \\
        \hline
        \multicolumn{5}{c}{\textbf{Low Frequency}} \\
        \hline
        SA-KLQR & 0.0412 & 0.0324 & 0.1421 & 0.0422 \\
        PID     & 0.0640 & 0.0502 & 0.2200 & 0.0654  \\
        SMC     & 0.0536 & 0.0421 & 0.1847 & 0.0548  \\
        \hline
        \multicolumn{5}{c}{\textbf{High Frequency}} \\
        \hline
        SA-KLQR & 0.0690 & 0.0524 & 0.1756 & 0.0450 \\
        PID     & 0.1070 & 0.0812 & 0.2712 & 0.0697  \\
        SMC     & 0.0897 & 0.0681 & 0.2283 & 0.0585  \\
        \hline
    \end{tabular}
\end{table}
\subsection{Tool Force Distribution and Coverage Evaluation}
To evaluate the effectiveness of the Centroid-Based Fuzzy Force Regulation approach, experiments were conducted with and without the proposed algorithm. The comparison focused on centroid error, force distribution uniformity, and task efficiency. The results, summarized in Table \ref{tab:centroid_comparison}, indicate that the algorithm significantly improves force distribution, reduces centroid error, and enhances task completion efficiency.

\begin{table}[h]
    \centering
    \caption{Performance Comparison With and Without the Centroid-Based Fuzzy Force Regulation Algorithm}
    \label{tab:centroid_comparison}
    \renewcommand{\arraystretch}{1.2}
    \setlength{\tabcolsep}{8pt} % Adjust column spacing
    \begin{tabular}{l|c|c}
        \hline
        \textbf{Test Condition} & \textbf{Standard} & \textbf{Centroid-Regulated} \\
        \hline
        \textbf{Average Centroid Error (cm)} & 2.5 & 0.8 \\
        \textbf{Coverage Percentage (\%)} & 85 & 97 \\
        \textbf{Average Force Error (\%)} & 10 & 3 \\
        \textbf{Task Completion Time (s)} & 83 & 95 \\
        \hline
    \end{tabular}
\end{table}

\begin{figure}
    \centering
    \includegraphics[width=1\linewidth]{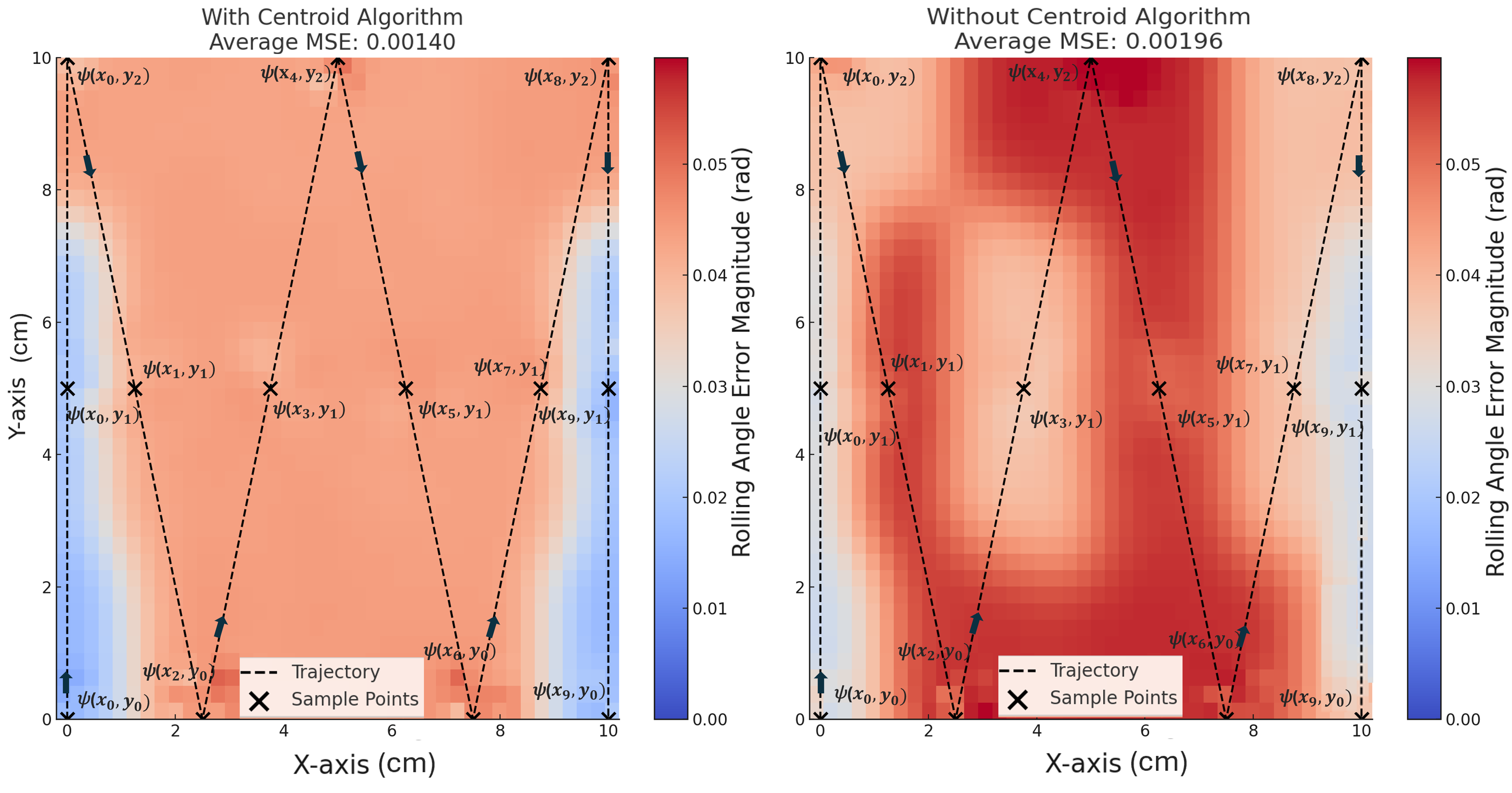}
    \caption{Rolling Angle Error with and Without Centroid Algorithm. The heatmaps show error distribution across the contact surface. (Left) With the algorithm, errors are lower and more uniform. (Right) Without it, errors concentrate at transitions, indicating instability.}
    \label{fig:heatmap_comparison}
\end{figure}

Figure \ref{fig:heatmap_comparison} presents heatmaps of the contact sensor pad, illustrating the difference in force distribution between the two cases. Without the algorithm, force application is inconsistent, leading to regions of excessive or insufficient pressure. In contrast, with the proposed approach, the force is more evenly distributed, ensuring effective contact without unnecessary rolling angle adjustments.

\begin{figure}[h]
    \centering
    \includegraphics[width=1\linewidth]{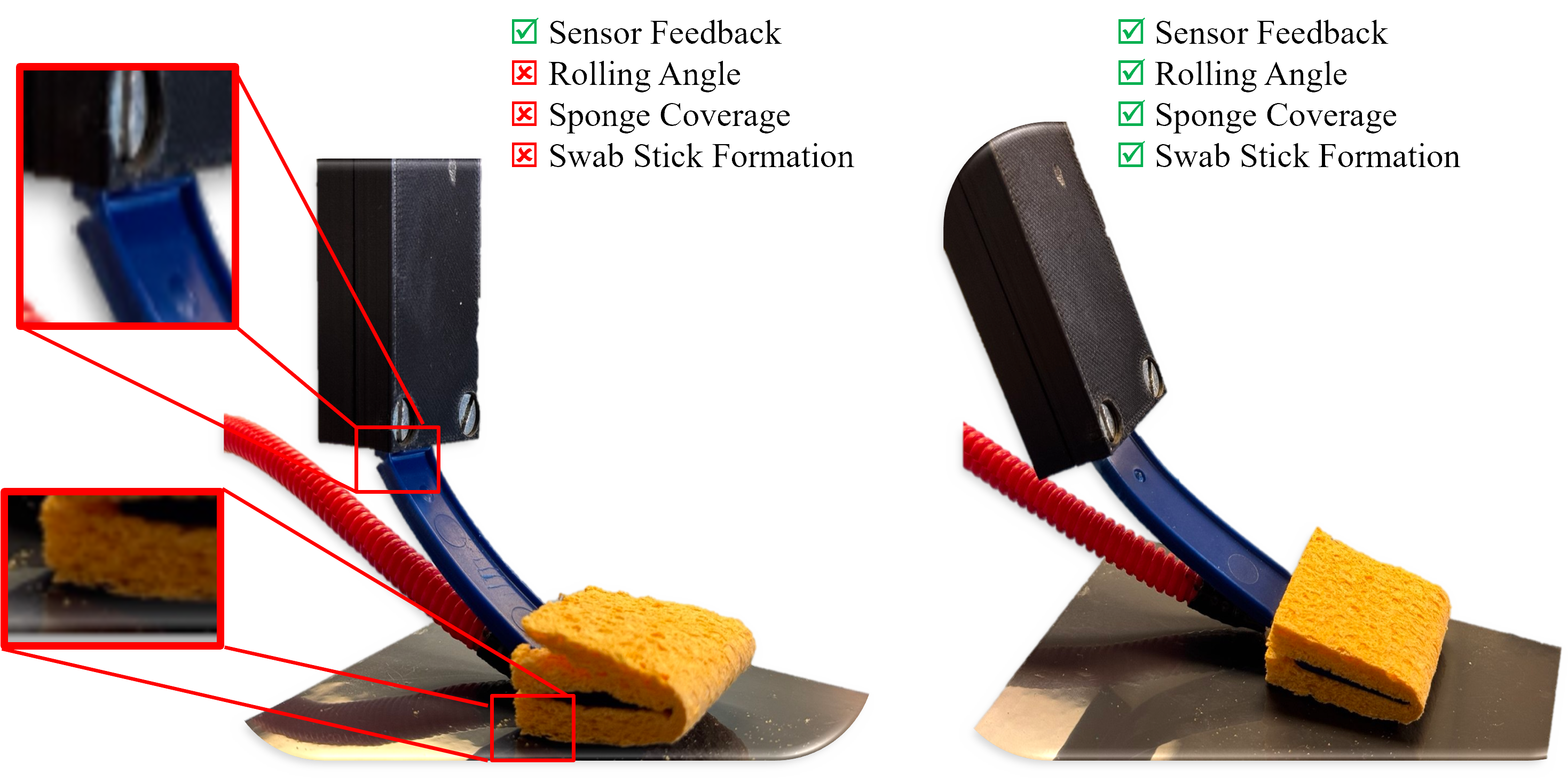}
    \caption{ Effect of Tool Distortion Due to Force Misalignment. (Left) Without the proposed algorithm, the controller compensates by increasing the rolling angle, causing excessive tool deformation. (Right) With the algorithm, force is evenly distributed, maintaining proper tool alignment and preventing distortion.}
    \label{fig:tool-distortion}
\end{figure}

A critical issue observed without the Centroid-Based Fuzzy Force Regulation algorithm is highlighted in Fig. \ref{fig:tool-distortion}. In some cases, the controller, unaware of tool deformation, increases the rolling angle excessively to meet the desired force, ultimately causing structural distortion of the tool. This uncontrolled compensation compromises both tool stability and effective force application. By integrating centroid-based feedback, the proposed algorithm dynamically corrects for these errors, preventing excessive deformations and maintaining stable force regulation throughout the swabbing process.
These findings underscore the necessity of force distribution monitoring in robotic swabbing applications, demonstrating that conventional force controllers alone are insufficient for handling compliant tools. The proposed centroid error fuzzy algorithm bridges this gap, ensuring robust and precise force control even in dynamic, flexible tool interactions.

\section{Future Work}
This research lays the foundation for advanced deformable tool manipulation, opening avenues for integrating learning-based approaches to enhance adaptability and precision. The developed framework has the potential to bridge the gap between conventional force control and fully autonomous learning-driven strategies, offering insights into how robots can achieve high-fidelity task execution. Additionally, leveraging system feedback and internal actuation patterns could provide alternative sensing methodologies, reducing dependence on external force measurement. Future explorations may focus on refining the system’s adaptability across varying conditions, improving learning efficiency, and expanding its applicability in complex industrial environments where automation can enhance both consistency and compliance.

\section{Conclusion}
In this paper, we proposed an adaptive Koopman-based force control framework for environmental swabbing in meat industries, addressing the challenges of deformable tool manipulation. By introducing SA-KLQR controller for precise force tracking and a centroid-based fuzzy algorithm for real-time compliance correction, the proposed approach ensures stable and efficient contact force regulation. Experimental results demonstrated that SA-KLQR outperformed conventional PID and SMC controllers, achieving lower tracking errors and improved stability across varying trajectory conditions. The centroid-based algorithm further enhanced force distribution, reducing tool distortion and ensuring consistent surface coverage. The findings highlight the effectiveness of the proposed method in automating environmental swabbing, improving accuracy, efficiency, and reliability in industrial hygiene applications.

% or


%\appendices

%\section{}

%\section*{Acknowledgment}


\ifCLASSOPTIONcaptionsoff
  \newpage
\fi



\bibliographystyle{IEEEtran}
\bibliography{bibtex/bib/IEEEexample}

\end{document}